\documentclass[11pt]{article}

\usepackage[final]{acl}

\usepackage{times}
\usepackage{latexsym}

\usepackage[T1]{fontenc}

\usepackage[utf8]{inputenc}

\usepackage{microtype}

\usepackage{inconsolata}

\usepackage{graphicx}

\usepackage{booktabs}
\usepackage{tabularx}
\usepackage{colortbl}
\usepackage{multirow}
\usepackage{amsmath}

\definecolor{forestgreen}{rgb}{0.13, 0.55, 0.13}
\definecolor{grey}{gray}{0.9}

\definecolor{worstscore}{RGB}{238,51,119} 
\definecolor{bestscore}{RGB}{51,187,238} 

\newcommand{\worst}[1]{\textcolor{worstscore}{#1}}
\newcommand{\best}[1]{\textcolor{bestscore}{#1}}
\newcommand\ciformat[1]{{\color{gray}\scriptsize$\pm$#1}}

\newcommand{\emldisplay}[2]{\texttt{\href{mailto:#1}{#2}}}
\newcommand{\eml}[1]{\emldisplay{#1}{#1}}

\title{You Have a Face for Radio: Bias in Multimodal Speech Recognition Models}

  \author{%
Maya K. Nachesa \qquad
Vlad Niculae \\
Language Technology Lab \\ University of Amsterdam\\
\{
\emldisplay{m.k.nachesa@uva.nl}{m.k.nachesa},~
\emldisplay{v.niculae@uva.nl}{v.niculae}
\}\texttt{@uva.nl}
\And
Vagrant Gautam  \\
Heidelberg Institute \\ for Theoretical Studies\\
\eml{vagrant.gautam@h-its.org}
}

\begin{document}
\maketitle

\begin{abstract}
As large neural models have become better at language tasks, researchers are increasingly building multi- and omnimodal models that handle more than one modality of data.
One example is the expansion of speech recognition models to audio-visual data, the goal being noise mitigation and multimodal subtitling.
While performance and bias have been studied extensively in the single-modality regime, it is unknown how new modalities affect this, even though they produce biases in humans.
We therefore propose the first bias evaluation of multimodal speech recognition, where we create
videos pairing different faces with the same audio using lip-sync,
and measure changes in
speech transcription accuracy.
We find large quality-of-service differences across \textsc{mWhisper-Flamingo} and \textsc{Gemini} models, with drops of up to $4.05$ word error rate points, across self-declared gender, ethnicity, and their intersection.
Our findings point to a priority for developers to evaluate, fix, and communicate such limitations,
as providing more signals through additional modalities is not necessarily better, and may even lead to biased outcomes.
\end{abstract}

\section{Introduction}

As large neural models improve on single  modalities of language data like text or audio, it is becoming increasingly common to incorporate multiple modalities into single ``multimodal'' or ``omnimodal'' models, such as images and text \citep{pmlr-v139-radford21a,NEURIPS2023_6dcf277e}, or video and audio \citep{avhubert, rouditchenko2024whisperflamingo, rouditchenko25_mwhisperflamingo, AV-LLM-video-understanding2025}.
Within audio-based research, the expansion to multimodality has focused on audio-visual speech recognition, with applications in noisy environments \citep{rouditchenko2024whisperflamingo, rouditchenko25_mwhisperflamingo}, automatic video captioning, subtitling \citep{2025DL-VC-review}, and more.

Auditory and visual processes are also implicated in human speech perception \citep{McGurk_MacDonald_1976}, where they similarly help us deal with noisy environments \citep{Sumby_Pollack_1954, vsabic2020noiseeyemovements}.
However, visual perceptions of the speaker can sometimes worsen our auditory processing of their speech, a phenomenon termed \textit{reverse linguistic stereotyping} \citep{kang-rubin-2009-reverse}.
In other words, nonlinguistic factors such as visually-perceived speaker ethnicity affect speech perception and processing \citep{Rubin_Smith_1990,Rubin_1992,Zheng_Samuel_2017,Bhuvanagiri_Grahlherr_Beardsell_Chittoor_Rubin_2019,Kutlu03072023,kutlu2022intelligibility}.
No study thus far has examined whether such multimodal raciolinguistic biases also appear in audio-visual speech recognition systems, despite a long history of study in single-modality settings in vision \citep{pmlr-v81-buolamwini18a} and in speech \citep{koenecke-et-al-2020-racial}.

The lack of such studies is largely due to the difficulty of collecting adequate data: Videos must be \emph{audio-matched} to eliminate confounding factors,
as in the matched-guise setup from sociolinguistics. 
In the matched-guise setup, the same speaker takes on different ``guises,'' which in our case corresponds to different faces.
Because the speakers remain the same, and only the guise is changed, it is ``the only possible factor that can influence people's evaluations'' \citep{stefanowitsch2005matchedguise}.
We propose a synthetic setup that allows us to evaluate whether the ethnicity and gender of different faces affects audio-visual speech recognition performance at scale.
We construct a dataset of 
$75,000$
videos where English audio samples that vary in auditory characteristics (accent and noise) are systematically paired with lip-synced faces that vary in visual characteristics (ethnicity and gender).
Using this dataset, we evaluate audio-visual speech recognition with two models: \textsc{mWhisper-Flamingo medium} \citep{rouditchenko25_mwhisperflamingo} and \textsc{Gemini-$2.5$-Flash} \citep{comanici2025gemini25pushingfrontier}.

We show that our synthetic setting is a valid testbed for bias, as our lip-synced videos help audio-visual speech recognition (AVSR) performance under noisy conditions, just like natural video.
Both with and without noise, AVSR models show single-axis and intersecting ethnicity and gender biases when transcribing the exact same set of audios with different faces.
Our findings thus provide evidence of reverse linguistic stereotyping in AVSR models.
We hope that our work spurs developers to fix bias in contexts where biases may come from multiple modalities, towards better and fairer multimodal speech recognition.\footnote{Code at: \href{https://github.com/ltl-uva/MultiModalSpeechBias}{github.com/ltl-uva/MultiModalSpeechBias}.}

\paragraph{Bias statement.}
In response to calls for clearer articulations of bias in language technology \citep{blodgett-etal-2020-language,wang-et-al-2022-measuring}, we conceptualize bias in our work as differences in quality-of-service with audio-visual speech recognition, based purely on a change in visual cues such as race and gender that should be irrelevant to transcription of the exact same audio.
This conceptualization mirrors that of \citet{harvey2025qualityofservice}, who ``present a framework for auditing LLM-based chatbots for dialect bias by measuring the extent to which they produce quality-of-service harms, which occur when systems do not work equally well for different people.''
Our definition of bias is parallel to reverse linguistic stereotyping in humans, and we measure it by testing for statistically significant changes in word error rate (WER), which can have detrimental effects on users of speech recognition technologies \citep{10.3389/frai.2021.725911,markl-2022-british-asr}.

\section{Related Works}

\paragraph{Multimodal speech perception in humans.}
Prior work in the field of raciolinguistics has shown that the exact same English audio gets different intelligibility ratings and scores from listeners when paired with pictures of White versus South Asian faces \citep{kutlu2022intelligibility,Kutlu03072023}, or White versus Asian faces \citep{Rubin_1992}.
However, this effect seems to disappear with video \citep{Zheng_Samuel_2017}.
Notably, these works use a very small sample size (three unambiguous words, presented on an accent continuum in \citealp{Zheng_Samuel_2017}), whereas we test at a much larger scale with automated and computational methods.
Our methods are inspired by this literature (in particular the matched-guise test setup), but we evaluate speech recognition systems rather than human listeners, and focus on transcription rather than speaker attitudes as in some of these studies.

\paragraph{Single-modality bias in vision and speech.}
Prior to our work, quality-of-service differentials based on race and gender have been shown in image classification \citep{pmlr-v81-buolamwini18a}, image captioning \citep{Hendricks_Burns_Saenko_Darrell_Rohrbach_2018,wolfe-caliskan-2022-markedness}, and English speech recognition \citep{tatman-2017-gender,koenecke-et-al-2020-racial}.
In contrast to our work, none of these papers study multi-modal speech recognition biases.
Additionally, work on bias in speech recognition tends to focus on acoustically-perceptible differences (accent, dialect, etc.), in contrast to our work, where identical audio is paired with different \textit{visual} input.

\paragraph{Multi- and omnimodal models.}
Analyses of multimodal models focus overwhelmingly on vision-language models that take images and text as input.
For instance, \citet{fu2025hidden} show that these models tend to overlook their vision input representations, \citet{kamruzzaman-etal-2025-seeing} analyze how race, gender, and skin tone affect these models' attribution of emotions, \citet{bui-etal-2025-multi3hate} evaluate multimodal, multicultural hate speech, and \citet{kim-etal-2025-tom} evaluate models' cultural stereotypes when visual ethnicity information is perturbed.
Although this last paper's methodology is similar to ours in terms of systematically perturbing ethnicity features, our work focuses on video and speech modalities, which none of the above consider.
Closest to our work, \citet{Dai_2024_CVPR} evaluate the robustness of audiovisual speech recognition to dropped frames, while we look at robustness to variance in the visual component of the input.


\section{Methodology}

The dominant signal for speech recognition is the audio.
In order to disentangle the impact of visual biases from audio-level performance differences in AVSR, we require a controlled setup with rich demographic information.
No such audio-visual dataset is publicly available.
Therefore, we create a usefully-close approximation through modifying and combining existing datasets.
The result allows us to compare the performance of AVSR models with varying visuals
(faces of varying ethnicities and genders) against the \emph{same} audio signal.

\subsection{Dataset}\label{sec:data}

To evaluate multimodal speech recognition bias, we create a dataset of videos that systematically combine faces and audio files, as described below.
This corresponds to experiments in the matched-guise setup with humans \citep{Zheng_Samuel_2017,kutlu2022intelligibility,Kutlu03072023}.

\paragraph{Faces.}
For high-resolution, standardized photographs of racially diverse faces, we use the Chicago Face Database \citep{Ma_Correll_Wittenbrink_2015} and the India Face Set \citep{lakshmi-et-al-cfd-india}.
From these datasets, we select a total of $500$ faces that these datasets mark as having a neutral expression.
We balance the faces equally across five ethnicities, with $100$ faces per ethnicity.
Using their terminology, these are: Asian American, Black, Indian Asian, Latino/a, and White.\footnote{In the rest of the text, we use the labels Asian, Black, Indian, Latinx, and White, for brevity.}
Within each ethnicity, we equally balance the faces we selected across Female and Male, the only two gender labels in the datasets. Thus, there are $250$ faces in each gender, and $50$ faces for any gender and ethnicity combination.
Ethnicity and gender in these datasets are self-reported, and we exclude faces of those who chose not to report.
We ensure that each individual's face is only included once.

\paragraph{Audio.}
We use English audio from the CommonVoice $17.0$ dataset \citep{ardila-etal-2020-common}, and select samples with accents that we will refer to as UK, US, and Indian.
We chose these accents, because they are also investigated by \citet{kutlu2022intelligibility} and \citet{Kutlu03072023}, and are well-represented in the dataset, meaning there were more higher-quality audios to choose from.
As accents in this dataset are self-identified, each grouping contains a mixed set of accent profiles, resulting in a sample that reflects some of the diversity of these coarse labels.
We describe accent selection in Appendix \ref{app:accent-audio-selection}.
Since all audio samples are self-recorded under diverse recording conditions, audio quality varies widely.
We use a heuristic based on audio ratings from CommonVoice to obtain the top $50$ audio samples per accent, as outlined in Appendix \ref{app:audio-sampling}.
We explicitly choose not to balance for gender, and in turn do not equally balance audio and video gender combinations, because the range of human voices is not strictly determined by gender:
Language and culture, socialization, intersectionality, and agency all affect how any voice sounds and is used \citep{zimman2018transgender}.
A model reacting to any perceived mismatches between audio and video would be ignoring this plurality of human ways of being, which would constitute another kind of bias.
Additionally, many participants do not indicate gender; ignoring the gender label thus also allows us to select recordings of higher quality.
With $50$ samples each from three accents, we have a total of $150$ distinct audio files.
As CommonVoice is not designed to collect the same text for different accents or speakers, there are no repeated texts.

\paragraph{Pairing faces and audio.}
All faces are combined with all audios, leading to a total of $500$ faces $\times$ $3$ accents $\times$ $50$ audios $=$ $75,000$ data points.
For each combination, we create an audio-video file where the video is lip-synced to the audio with \textsc{Wav2Lip $+$ GAN} \citep{wav2lip}, using the clean audio data as input.
All videos are first downscaled to a height of $256$, maintaining the original aspect ratio, to fit the lip-sync model.
Our use of lip-synced videos reflects realistic and intended usage of audio-visual models, in contrast to most sociolinguistic work where still faces are used.
We performed a manual check on a subset of the videos to verify that the lip-synced videos looked correct.
We additionally provide results with videos of static faces in Appendix \ref{app:still-faces}, for reference.

\begin{table}[t]
    \centering
    \small
    \begin{tabularx}{\linewidth}{Xlrr}
        \toprule
        \textbf{Model} & \textbf{Accent} &
        \textbf{Average} &
        \textbf{ASR}
        \\
        \midrule
        \rowcolor{grey} \multicolumn{4}{c}{No noise (SNR$ = \infty$)} \\
\textsc{mWhisper-F} & Indian   &
24.13\ciformat{.43}	&21.53\ciformat{7.33\phantom{8}}\\
\textsc{mWhisper-F} & US       &
13.47\ciformat{.25}	&12.94\ciformat{4.72\phantom{8}}\\
\textsc{mWhisper-F} & UK	   &
19.78\ciformat{.35}	&17.77\ciformat{6.90\phantom{8}}\\

\midrule
\textsc{Gemini} & Indian & 
10.84\ciformat{.24} & 7.24\ciformat{3.13\phantom{8}} \\ 
\textsc{Gemini} & US & 
8.43\ciformat{.18} & 6.67\ciformat{3.60\phantom{8}} \\ 
\textsc{Gemini} & UK &
11.73\ciformat{.28} & 8.78\ciformat{4.12\phantom{8}} \\ 
\midrule
\rowcolor{grey} \multicolumn{4}{c}{Babble noise (SNR$=5$ dB)} \\
\textsc{mWhisper-F} & Indian   &
58.31\ciformat{.52}	&49.90\ciformat{9.63\phantom{8}}\\
\textsc{mWhisper-F} & US	   &
34.75\ciformat{.43}	&31.76\ciformat{8.10\phantom{8}}\\
\textsc{mWhisper-F} & UK	   &
46.49\ciformat{.54}	&47.97\ciformat{15.64}\\
\midrule
\textsc{Gemini} & Indian & 
39.16\ciformat{.53} & 43.05\ciformat{11.61} \\
\textsc{Gemini} & US &
23.60\ciformat{.31} & 27.25\ciformat{7.96\phantom{8}} \\
\textsc{Gemini} & UK & 
34.29\ciformat{.49} & 34.26\ciformat{9.44\phantom{8}} \\
        \bottomrule
    \end{tabularx}
    \caption{Average per-accent WER scores of AVSR systems compared to audio-only ASR versions thereof. 
    \textsc{mWhisper-F} = \textsc{mWhisper-Flamingo medium}; \textsc{Gemini} = \textsc{Gemini-$2.5$-flash}.}
    \label{tab:wer-scores}
\end{table}

Prior work \citep{rouditchenko2024whisperflamingo, rouditchenko25_mwhisperflamingo} has shown that AVSR mostly benefits from visual input in the presence of noise, with no benefits under no noise. Thus, we also generate noisy versions of the audio-video files by adding ``babble'' noise with a Signal-to-Noise Ratio (SNR) of $5$ dB to the audio track.
We obtain this noise from \citet{rouditchenko2024whisperflamingo}, which in turn is sampled from the MUSAN dataset \citep{snyder2015musan}.

\paragraph{Lip-sync as a useful approximation.}
A matched-guise setup is a balance between precision and scale.
One the one hand, real videos visually match what models most likely face in practice, but require quite an intricate setup. 
For example, \citet{Zheng_Samuel_2017} used a total of three single-word stimuli that were recorded multiple times by four speakers, two for the audio, and two for the video, so that in the end the ones with the same speaking rate could be chosen.
However, to expand this setup (such that there are more individuals in the audio and video recordings, and more and longer stimuli) would thus constitute a considerable effort.

On the other hand, using lip-sync allows our sample audios to be longer, as well as for more varied samples, speakers, and faces and ethnicity to be associated with any one audio.
Moreover, \citet{shan2022aiVSrealVidNoiseComp} show that, when reducing the SNR from $-9$ to $0$ dB\footnote{$\text{SNR}>0$ dB means more signal than noise; $\text{SNR}<0$ dB means more noise than signal.}, participants' accuracy for real and synthesized videos improved equally compared to the audio-only setting; our noisy setting with an SNR of 5 dB could fall within the same pattern.
Nevertheless, we recognize that the distribution of real and synthesized videos may not be the same.
However, we can still study the patterns that we observe in synthesized videos.

\begin{table*}[t]
    \centering
    \small
    \begin{tabularx}{\linewidth}{Xlrr@{}c@{\,}c@{\,}c@{}lrr@{}c@{\,}c@{\,}c@{}lrr@{}l@{\:}l@{\,}c@{\,}l@{\:}l@{}l}
        \toprule
        \multirow{2}{*}{\textbf{Model}} & \multirow{2}{*}{\textbf{Accent}} & \multicolumn{20}{c}{\textbf{$\Delta$ WER (\textbf{\worst{Max}}$-$\textbf{\best{Min}})}} \\
         & &
        \multicolumn{6}{c}{\textbf{Ethnicity}} &
        \multicolumn{6}{c}{\textbf{Gender}} &
        \multicolumn{8}{c}{\textbf{Intersection}}
        \\
        \midrule
        \rowcolor{grey} \multicolumn{22}{c}{No noise (SNR$ = \infty$)} \\
        \textsc{mWhisper-F} & Indian & $2.06^{\phantom{*}}$ & ( & \worst{Indian} & $-$ & \best{Asian} & ) & $1.14^{\phantom{*}}$          & ( & \worst{M} & $-$ & \best{F} & ) & $4.05^*$             & ( & \worst{Indian} & \worst{male}   & $-$ & \best{Asian}  & \best{male}   & ) \\
        \textsc{mWhisper-F} & US     & $0.72^{\phantom{*}}$ & ( & \worst{Black}  & $-$ & \best{Asian} & ) & $0.43^{\phantom{*}}$          & ( & \worst{M} & $-$ & \best{F} & ) & $0.98^{\phantom{*}}$ & ( & \worst{Black}  & \worst{male}   & $-$ & \best{Latinx} & \best{female} & ) \\
        \textsc{mWhisper-F} & UK     & $0.51^{\phantom{*}}$ & ( & \worst{Indian} & $-$ & \best{Asian} & ) & $1.34^*$                      & ( & \worst{M} & $-$ & \best{F} & ) & $2.27^*$             & ( & \worst{Indian} & \worst{male}   & $-$ & \best{Latinx} & \best{female} & ) \\
        \midrule
        \textsc{Gemini}     & Indian & $0.77^{\phantom{*}}$ & ( & \worst{Indian} & $-$ & \best{White} & ) & \textbf{$0.25^{\phantom{*}}$} & ( & \worst{F} & $-$ & \best{M} & ) & $1.19^*$             & ( & \worst{Indian} & \worst{male}   & $-$ & \best{White}  & \best{male}   & ) \\
        \textsc{Gemini}     & US     & $0.13^{\phantom{*}}$ & ( & \worst{Indian} & $-$ & \best{White} & ) & \textbf{$0.08^{\phantom{*}}$} & ( & \worst{M} & $-$ & \best{F} & ) & $0.26^{\phantom{*}}$ & ( & \worst{Asian}  & \worst{male}   & $-$ & \best{Asian}  & \best{female} & ) \\
        \textsc{Gemini}     & UK     & $0.49^{\phantom{*}}$ & ( & \worst{Indian} & $-$ & \best{Black} & ) & \textbf{$0.35^{\phantom{*}}$} & ( & \worst{M} & $-$ & \best{F} & ) & $0.90^{\phantom{*}}$ & ( & \worst{Indian} & \worst{male}   & $-$ & \best{Black}  & \best{female} & ) \\
        \midrule
        \rowcolor{grey} \multicolumn{22}{c}{Babble noise (SNR$ = 5$ dB)} \\
        \textsc{mWhisper-F} & Indian & $1.01^{\phantom{*}}$ & ( & \worst{Indian} & $-$ & \best{Asian} & ) & $0.10^{\phantom{*}}$          & ( & \worst{F} & $-$ & \best{M} & ) & $1.71^{\phantom{*}}$ & ( & \worst{Indian} & \worst{male}   & $-$ & \best{Asian}  & \best{male}   & ) \\
        \textsc{mWhisper-F} & US     & $1.75^*$             & ( & \worst{Indian} & $-$ & \best{Asian} & ) & $1.45^*$                      & ( & \worst{M} & $-$ & \best{F} & ) & $3.95^*$             & ( & \worst{Indian} & \worst{male}   & $-$ & \best{Asian}  & \best{female} & ) \\
        \textsc{mWhisper-F} & UK     & $0.91^{\phantom{*}}$ & ( & \worst{Black}  & $-$ & \best{White} & ) & $0.47^{\phantom{*}}$          & ( & \worst{F} & $-$ & \best{M} & ) & $2.84^*$             & ( & \worst{Indian} & \worst{male}   & $-$ & \best{Latinx} & \best{male}   & ) \\
        \midrule
        \textsc{Gemini}     & Indian & $1.49^*$             & ( & \worst{Asian}  & $-$ & \best{White} & ) & \textbf{$0.51^{\phantom{*}}$} & ( & \worst{M} & $-$ & \best{F} & ) & $2.73^*$             & ( & \worst{Asian}  & \worst{male}   & $-$ & \best{White}  & \best{male}   & ) \\
        \textsc{Gemini}     & US     & $0.53^{\phantom{*}}$ & ( & \worst{Black}  & $-$ & \best{White} & ) & \textbf{$0.20^{\phantom{*}}$} & ( & \worst{F} & $-$ & \best{M} & ) & $0.95^*$             & ( & \worst{Latinx} & \worst{female} & $-$ & \best{White}  & \best{male}   & ) \\
        \textsc{Gemini}     & UK     & $1.03^{\phantom{*}}$ & ( & \worst{Black}  & $-$ & \best{White} & ) & \textbf{$0.43^{\phantom{*}}$} & ( & \worst{M} & $-$ & \best{F} & ) & $1.62^*$             & ( & \worst{Black}  & \worst{male}   & $-$ & \best{White}  & \best{female} & ) \\
        \bottomrule
    \end{tabularx}
    \caption{Largest AVSR quality-of-service differences between faces, across ethnicity, gender, and their intersection. Asterisk (*) denotes statistical significance ($p < 0.005$) according to
    a permutation test (Section \ref{sec:expsetup}). \textsc{mWhisper-F} = \textsc{mWhisper-Flamingo medium}; \textsc{Gemini} = \textsc{Gemini-$2.5$-flash}. M = Male; F = Female.}
    \label{tab:quality-of-service-differences}
\end{table*}

\subsection{Experimental Setup}\label{sec:expsetup}

\paragraph{Models.}
We ran our experiments using \textsc{mWhisper-Flamingo medium}\footnote{We ran preliminary experiments with \textsc{mWhisper-Flamingo small} and \textsc{Whisper-Flamingo}, but exclude those as they perform much worse.} \citep{rouditchenko25_mwhisperflamingo} and \textsc{Gemini-$2.5$-Flash} \citep{comanici2025gemini25pushingfrontier}. As input, \textsc{mWhisper-Flamingo} takes audio and greyscaled video, cropped to the mouth region, separately.
As for \textsc{Gemini}, the full video is encoded in Base64.
We show prompts for \textsc{Gemini} in Appendix \ref{app:gemini-prompt}. We also report results with an automatic speech recognition (ASR) baseline obtained by turning off the video input, which both models support.
\textsc{mWhisper-Flamingo medium} was chosen as a SOTA representative of AVSR, lipreading-specific models, while \textsc{Gemini-$2.5$-Flash} was selected as a representative of API-based multi-modal LLMs, which, due to its generic nature, may also be used for video transcription, in addition to other tasks that a user may engage it for.

\paragraph{Metrics.}

We assess model performance with WER: The corpus-level 
average number of errors weighted by document length.
Implementation notes are provided in Appendix \ref{app:wer-calculation}.
We also compute WER standard deviations as a weighted average of the per-document difference to the mean,
and estimate standard errors of the mean WER using the normal assumption.

For statistical testing of significant pairwise quality-of-service differences (Table \ref{tab:quality-of-service-differences}), we apply $10,000$ iterations of a one-sided permutation test, stratified over audio files. To correct for 
multiple comparisons within each row, we apply the \emph{max-T} method \citep{maxt}: We construct the permutation sampling distribution of maximum pairwise differences $\max_{u,v \in C}|\text{WER}_{u} - \text{WER}_{v}|$,
where $C$ is the set of possible treatment conditions, including
marginal ones (e.g., \texttt{gender=male}, \texttt{ethnicity=asian})
and intersections (e.g., \texttt{gender=male and ethnicity=asian}).
Within each table row, all conditions of interest are thus comparable against this same sampling distribution. To correct for the number of rows, we use the Bonferroni method.

\section{Main Findings}

Table \ref{tab:wer-scores} provides an overview of WERs on average, compared to an ASR-only (i.e., no visual input) baseline. See Appendix \ref{app:full-results} for a full breakdown by gender and ethnicity conditions.
Overall, \textsc{Gemini} models perform better than \textsc{mWhisper-Flamingo} models in all settings (despite some hallucinations and deviations from instructions - see Appendix \ref{app:HR-the-fun-kind}).
We caution, however, that \textsc{Gemini}'s better performance could be due to train-test contamination.
US accents are transcribed better than UK accents, which in turn are transcribed better than Indian accents.
These performance differences could also come from differences in audio quality, however, they pattern similarly to previously-studied accent biases in ASR models \citep{chan22b_interspeech,dichristofano2023globalperformancedisparitiesenglishlanguage}.

\paragraph{Lip-synced input is useful for AVSR under noise.}
In the absence of noise, visual input degrades recognition performance. 
However, when we add noise at an SNR of $5$ dB, compared to ASR alone, \textsc{mWhisper-Flamingo} improves for UK accents, and \textsc{Gemini} improves in all settings.
This aligns with the motivation and results in \citet{rouditchenko2024whisperflamingo} and \citet{rouditchenko25_mwhisperflamingo}, 
%
and suggests that our synthetic data setup
resembles realistic conditions.
Audio signals are often enough for ASR, as they are in some sense captured more faithfully than video (where mouths may be badly cropped or undetected, or faces may be off-camera).
Nevertheless, the fact that video sometimes \textit{worsens} rather than maintains performance under no noise points to the infancy of these models.

\paragraph{Models show ethnicity and gender biases when transcribing the same audio with different faces.}

As Table \ref{tab:quality-of-service-differences} shows, there are several statistically significant quality-of-service differences in transcribing the exact same set of audios with different groupings of faces, across both models and all accent classes.
While some single-axis ethnicity and gender differences may not be significant, they hide \textit{intersecting} biases as in other settings in both vision \citep{pmlr-v81-buolamwini18a} and language \citep{rodriguez-etal-2025-colombian}:
Under no noise, Indian male faces consistently worsen \textsc{mWhisper-Flamingo}'s transcription, and White male faces consistently get the best WERs with \textsc{Gemini}.
In fact, results even seem to worsen under noise, where we have seen that models rely more on their visual input as well.
These results provide evidence of the reverse linguistic stereotyping effect in models, and point to an important priority for model developers to fix.




\paragraph{Bias unrelated to lip-sync.}
To support these findings, we also analyze the effect on static faces. Table \ref{tab:quality-of-service-differences-static} in Appendix \ref{app:still-faces} shows that, even with static faces, both \textsc{mWhisper-Flamingo} and \textsc{Gemini} sometimes show quality-of-service differences based on ethnicity, gender, or their intersection.
Thus, the bias observed in this setting comes from the AVSR models alone, and is unrelated to lip-sync.
This, together with the observation that different groups experience the largest quality-of-service difference after lip-sync, but that this change is not consistent, supports the conclusion that AVSR models react differently to the same audio when presented with different faces.

\section{Conclusion}
We contribute a framework to study biases in audiovisual speech recognition by creating videos that present the same audio with lip-synced faces that vary along the dimensions of ethnicity and gender.
WERs with \textsc{mWhisper-Flamingo} and \textsc{Gemini-$2.5$-flash} confirm that our synthetic setup, while inferior to real videos of people speaking, simulates it well enough, as it improves transcription over audio-only ASR under noise.
We then show that models show single-axis and intersecting ethnicity and gender biases when transcribing the same audio with different faces.
Our results provide evidence of reverse linguistic stereotyping in AVSR models, and suggest that synthetic data with lip-sync might be a useful avenue to mitigate these quality-of-service differences.
Fixing such biases that emerge from the combination of multiple modalities is an important priority for the future of multi- and omnimodal language technology and research.


\section*{Limitations}
The primary limitation of this work is that we evaluate with stand-ins for real video: Lip-synced videos in the main text, and videos with static faces in Appendix \ref{app:still-faces}.
Lip-sync may introduce artifacts or mistakes compared to real videos, such as movements when there is no speech or vice versa, the wrong movement for a certain sequence of phonemes, or in general produce output ranging from homogenized or not-quite-realistic mouth shapes up to unusable or fully noisy outputs.
The model we used only animates the mouth portion of the face, meaning movements in the rest of the face, the jaw, and general head positioning may not relate to the content of the speech. 
Finally, there may be other (invisible) artifacts that may separate lip-synced videos from real videos for a video model.
The second limitation is that we work with a limited set of accents and faces, as well as only in English, while results are sure to differ in other languages and with a broader variety of facial and dialectal differences.
Finally, we use self-declared ethnicity, gender, and accent labels, which is good for participant autonomy and dignity, but we note that \textit{perceptions} of people may differ from identity, and are thus still likely to play a larger role in bias in computational systems.

\section*{Ethics Statement}

Ethnicity and gender are multi-faceted aspects of identity, and our operationalization choices affect the conclusions we can draw \citep{steidl-werum-2019-operationalization}.
Ethnicity, gender, and accent are all self-declared in the data we use.
Although this supports participant autonomy and dignity, perceptions and behaviour (from both humans and machines) may differ, which can cause harm to individuals.
Our results show quality-of-service differentials in AVSR that replicate structural power differentials in society and are an example of algorithmic oppression, with harms to users of these technologies with certain faces.
We use all datasets in compliance with their terms of use and licensing restrictions, i.e., we do not identify the owners of the faces or voices, we use the face data only for non-commercial scientific research, and we use CommonVoice for evaluating speech recognition models.

\section*{Acknowledgments}
We thank Aashutosh Ganesh, Annika Kniele, Daniella Pauly Jensen, the (D)NN-speech reading group at the University of Amsterdam, and the reviewers for their valuable feedback.
We are grateful to RTL Netherlands for compute resources.
This publication is part of the project \emph{ROBUST: Trustworthy AI-based Systems for Sustainable Growth}
with project number KICH3.LTP.20.006, which is (partly) financed by the Dutch Research Council
(NWO), RTL Netherlands, and the Dutch Ministry of Economic Affairs and Climate Policy (EZK) under the program
LTP KIC 2020-2023. 
Vagrant Gautam's work is funded by the Klaus Tschira Foundation, Heidelberg, Germany.
Vlad Niculae is supported by the Dutch Research Council (NWO) via VI.Veni.212.228.

\bibliography{sample}

\appendix
\section{Accents in CommonVoice}
\label{app:accent-audio-selection}

CommonVoice includes self-declared accent labels.
First, we lowercase all labels.
All audio files that include ``india'' in the label, and that exclude ``united states'' and ``england english'' are selected as Indian accented audio files for our dataset.
We select all files that include ``england english'' in the label as UK-accented data.
Finally, all labels that include ``united states english'' are counted as US accents.
Through manually inspecting the remaining labels, we confirmed that no other accents were declared.

\section{Quality Filtering for CommonVoice}
\label{app:audio-sampling}

CommonVoice data includes audio ratings consisting of up-votes and down-votes.
We designed a custom metric to use these ratings to select audios with the highest quality. 

For a given accent, we first weight the up-votes and down-votes. 
We define $U_i$ as the up-votes for sample $i$ and $U_\text{max}$ as the max up-votes for the accent.
Equivalently, we define $D_i$ as the down-votes for sample $i$ and $D_\text{max}$ as the max down-votes for the accent. 

Finally, we define $\bar{U}_i$ as the weighted up-votes for a given sample, and $\bar{D}_i$ as the weighted down-votes for a given sample, which we calculate as follows:
\begin{equation}
    \bar{U}_i := \frac{U_i}{U_\text{max}}
\end{equation}
\begin{equation}
    \bar{D}_i := \frac{D_\text{max} - D_i}{D_\text{max}}.
\end{equation}

A higher weighted score represents more up-votes and fewer down-votes, respectively. 

We then compute the score of an audio as the harmonic mean of the weighted votes, and take the top $50$ with the highest score per accent.

\begin{equation}
    \text{Score} := \frac{2\cdot \bar{U}_i \cdot \bar{D}_i}{\bar{U}_i + \bar{D}_i}.
\end{equation}

\begin{table}[t]
    \centering
    \small
    \begin{tabularx}{\linewidth}{Xlrr}
        \toprule
        \textbf{Model} & \textbf{Accent} &
        \textbf{Average} &
        \textbf{ASR}
        \\
        \midrule
        \rowcolor{grey} \multicolumn{4}{c}{No noise (SNR$ = \infty$)} \\
\textsc{mWhisper-F} & Indian   &
28.26\ciformat{.48\phantom{8}}	&21.53\ciformat{7.33\phantom{8}}\\
\textsc{mWhisper-F} & US       &
14.23\ciformat{.25\phantom{8}}	&12.94\ciformat{4.72\phantom{8}}\\
\textsc{mWhisper-F} & UK	   &
19.65\ciformat{.34\phantom{8}}	&17.77\ciformat{6.90\phantom{8}}\\

\midrule
\textsc{Gemini} & Indian & 
9.84\ciformat{0.19\phantom{8}} & 7.24\ciformat{3.13\phantom{8}} \\ 
\textsc{Gemini} & US & 
8.21\ciformat{0.17\phantom{8}} & 6.67\ciformat{3.60\phantom{8}} \\ 
\textsc{Gemini} & UK &
9.85\ciformat{0.22\phantom{8}} & 8.78\ciformat{4.12\phantom{8}} \\ 
\midrule
\rowcolor{grey} \multicolumn{4}{c}{Babble noise (SNR$=5$ dB)} \\
\textsc{mWhisper-F} & Indian   &
70.98\ciformat{1.28\phantom{8}}	&49.90\ciformat{9.63\phantom{8}}\\
\textsc{mWhisper-F} & US	   &
40.43\ciformat{0.85\phantom{8}}	&31.76\ciformat{8.10\phantom{8}}\\
\textsc{mWhisper-F} & UK	   &
49.92\ciformat{0.53\phantom{8}}	&47.97\ciformat{15.64}\\
\midrule
\textsc{Gemini} & Indian & 
37.97\ciformat{0.47\phantom{8}} & 43.05\ciformat{11.61} \\
\textsc{Gemini} & US &
24.06\ciformat{1.02\phantom{8}} & 27.25\ciformat{7.96\phantom{8}} \\
\textsc{Gemini} & UK & 
41.93\ciformat{14.21} & 34.26\ciformat{9.44\phantom{8}} \\
        \bottomrule
    \end{tabularx}
    \caption{Static-face (no lip-sync) absolute WER scores by groupings of faces across gender, ethnicity, and accent. \textsc{mWhisper-F} = \textsc{mWhisper-Flamingo medium}; \textsc{Gemini} = \textsc{Gemini-$2.5$-flash}.}
    \label{tab:wer-scores-static}
\end{table}

\section{Results with Static Faces}
\label{app:still-faces}
Tables \ref{tab:wer-scores-static} and \ref{tab:quality-of-service-differences-static} provide an overview of the quality-of-service differences for static faces.

\begin{table*}[t]
    \centering
    \small
    \begin{tabularx}{\linewidth}{Xl rr@{}c@{\,}c@{\,}c@{}l rr@{}c@{\,}c@{\,}c@{}l rr@{}l@{\:}l@{\,}c@{\,}l@{\:}l@{}l}
        \toprule
        \multirow{2}{*}{\textbf{Model}} & \multirow{2}{*}{\textbf{Accent}} & \multicolumn{20}{c}{\textbf{$\Delta$ WER (\textbf{\worst{Max}}$-$\textbf{\best{Min}})}} \\
         & &
        \multicolumn{6}{c}{\textbf{Ethnicity}} &
        \multicolumn{6}{c}{\textbf{Gender}} &
        \multicolumn{8}{c}{\textbf{Intersection}}
        \\
        \midrule
        \rowcolor{grey} \multicolumn{22}{c}{No noise (SNR$ = \infty$)} \\
        \textsc{mWF}    & Indian & $3.22^{        {*}}$ & ( & \worst{Latinx} & $-$ & \best{Indian} & ) &         $0.36^{\phantom{*}}$  & ( & \worst{M} & $-$ & \best{F} & ) &  $4.81^{        {*}}$ & ( & \worst{White}  & \worst{female} & $-$ & \best{Indian} & \best{female} & ) \\
        \textsc{mWF}    & US     & $0.82^{\phantom{*}}$ & ( & \worst{Latinx} & $-$ & \best{Indian} & ) &         $0.22^{\phantom{*}}$  & ( & \worst{M} & $-$ & \best{F} & ) &  $1.46^{        {*}}$ & ( & \worst{Latinx} & \worst{male}   & $-$ & \best{Indian} & \best{female} & ) \\
        \textsc{mWF}    & UK     & $0.58^{\phantom{*}}$ & ( & \worst{Asian}  & $-$ & \best{Indian} & ) &         $0.15^{\phantom{*}}$  & ( & \worst{M} & $-$ & \best{F} & ) &  $0.87^{\phantom{*}}$ & ( & \worst{Asian}  & \worst{male}   & $-$ & \best{Indian} & \best{male}   & ) \\
        \midrule
        \textsc{Gemini} & Indian & $0.41^{\phantom{*}}$ & ( & \worst{Indian} & $-$ & \best{White}  & ) & \textbf{$0.37^{\phantom{*}}$} & ( & \worst{F} & $-$ & \best{M} & ) &  $0.77^{\phantom{*}}$ & ( & \worst{Latinx} & \worst{female} & $-$ & \best{White}  & \best{male}   & ) \\
        \textsc{Gemini} & US     & $0.17^{\phantom{*}}$ & ( & \worst{Black}  & $-$ & \best{White}  & ) & \textbf{$0.00^{\phantom{*}}$} & ( & \multicolumn{3}{c}{n/a}  & ) &  $0.30^{\phantom{*}}$ & ( & \worst{Indian} & \worst{female} & $-$ & \best{Indian} & \best{male}   & ) \\
        \textsc{Gemini} & UK     & $0.31^{\phantom{*}}$ & ( & \worst{Asian}  & $-$ & \best{White}  & ) & \textbf{$0.19^{\phantom{*}}$} & ( & \worst{F} & $-$ & \best{M} & ) &  $0.57^{\phantom{*}}$ & ( & \worst{Asian}  & \worst{female} & $-$ & \best{White}  & \best{female} & ) \\
        \midrule
        \rowcolor{grey} \multicolumn{22}{c}{Babble noise (SNR$ = 5$)} \\
        \textsc{mWF}    & Indian & $8.46^{        {*}}$ & ( & \worst{White}  & $-$ & \best{Black}  & ) &         $0.54^{\phantom{*}}$  & ( & \worst{M} & $-$ & \best{F} & ) & $12.01^*$             & ( & \worst{White}  & \worst{male}   & $-$ & \best{Black}  & \best{male}   & ) \\
        \textsc{mWF}    & US     & $1.75^{        {*}}$ & ( & \worst{Latinx} & $-$ & \best{Asian}  & ) &         $0.18^{\phantom{*}}$  & ( & \worst{F} & $-$ & \best{M} & ) &  $5.29^*$             & ( & \worst{Black}  & \worst{female} & $-$ & \best{Black}  & \best{male}   & ) \\
        \textsc{mWF}    & UK     & $2.12^{        {*}}$ & ( & \worst{Latinx} & $-$ & \best{Indian} & ) &         $0.00^{\phantom{*}}$  & ( & \multicolumn{3}{c}{n/a}  & ) &  $3.07^*$             & ( & \worst{Latinx} & \worst{female} & $-$ & \best{Indian} & \best{female} & ) \\
        \midrule
        \textsc{Gemini} & Indian & $0.55^{\phantom{*}}$ & ( & \worst{Latinx} & $-$ & \best{Indian} & ) &         $0.15^{\phantom{*}}$  & ( & \worst{M} & $-$ & \best{F} & ) &  $0.77^{\phantom{*}}$ & ( & \worst{Latinx} & \worst{male}   & $-$ & \best{Indian} & \best{female} & ) \\
        \textsc{Gemini} & US     & $3.31^{        {*}}$ & ( & \worst{Asian}  & $-$ & \best{White}  & ) &         $1.35^{        {*}}$  & ( & \worst{F} & $-$ & \best{M} & ) &  $6.22^*$             & ( & \worst{Asian}  & \worst{female} & $-$ & \best{White}  & \best{female} & ) \\
        \textsc{Gemini} & UK     & $31.18^{       {*}}$ & ( & \worst{Asian}  & $-$ & \best{White}  & ) &        $13.16^{        {*}}$  & ( & \worst{F} & $-$ & \best{M} & ) & $63.40^*$             & ( & \worst{Asian}  & \worst{female} & $-$ & \best{Asian}  & \best{male}   & ) \\
        \bottomrule
    \end{tabularx}
    \caption{Static-face (no lip-sync) largest AVSR quality-of-service differences between faces, across ethnicity, gender, and their intersection. Asterisk (*) denotes statistical significance ($p < 0.005$) according to
    a permutation test (Section \ref{sec:expsetup}). \textsc{mWF} = \textsc{mWhisper-Flamingo medium}; \textsc{Gemini} = \textsc{Gemini-$2.5$-flash}. M = Male; F = Female.}
    \label{tab:quality-of-service-differences-static}
\end{table*}

\section{Prompting \textsc{Gemini}}
\label{app:gemini-prompt}

We pass two text prompts to \textsc{Gemini} together with the video or audio: A system prompt and an instruction.
The system prompt was ``You are a helpful transcription agent.''

The instruction for video transcription was ``Please transcribe everything that is said in the following video into English. Output only the transcription. Prioritize accuracy in transcription, including punctuation and sentence structure, to reflect natural speech flow. The audio quality may vary from good to quite noisy, please do your best to isolate the main speaker.'' For audio transcription, we replace the word ``video'' with ``audio''.

\section{Word Error Rate Calculations}
\label{app:wer-calculation}

We calculate Word Error Rate with the Fairseq \citep{fairseq} \texttt{WerScorer}, which we initialize as follows:

\noindent\texttt{scorer = WerScorer(}

\texttt{WerScorerConfig(wer\_tokenizer="13a",}

\texttt{wer\_remove\_punct=True,}

\texttt{wer\_char\_level=False,}

\texttt{wer\_lowercase=True))}



\section{Full Results}\label{app:full-results}

All WER results with confidence intervals across all splits are shown in Table \ref{tab:all-results}.

\section{Hallucinations and Refusals}\label{app:HR-the-fun-kind}

Here, we provide some examples of model hallucinations (output unrelated to input) in transcriptions, (partial) refusals to transcribe, and other unrequested output.

Table \ref{tab:gemini-hallucinations} provides an example of a hallucination.

\begin{table*}[ht]\centering\small
    \begin{tabular}{llrrrrrrr}
    \toprule
    \textbf{Model}             & \textbf{Accent} & \textbf{Gender} & \textbf{Asian} & \textbf{Black} & \textbf{Indian} & \textbf{Latinx} & \textbf{White} & \textbf{Average} \\
    \midrule
    \textsc{mWhisper-F} & Indian &  Female & 23.30\ciformat{1.28} & 23.97\ciformat{1.24} & 23.37\ciformat{0.99} & 23.68\ciformat{1.17} & 23.46\ciformat{1.02} & 23.56\ciformat{0.51} \\
                    &        &    Male & 22.97\ciformat{1.05} & 24.85\ciformat{1.33} & 27.03\ciformat{1.77} & 23.80\ciformat{1.05} & 24.83\ciformat{2.14} & 24.70\ciformat{0.68} \\
                    &        & Average & 23.14\ciformat{0.83} & 24.41\ciformat{0.91} & 25.20\ciformat{1.01} & 23.74\ciformat{0.79} & 24.14\ciformat{1.18} & 24.13\ciformat{0.43} \\
    
    \textsc{mWhisper-F} &     US &  Female & 13.04\ciformat{0.67} & 13.91\ciformat{1.07} & 13.07\ciformat{0.68} & 13.01\ciformat{0.67} & 13.25\ciformat{0.69} & 13.26\ciformat{0.34} \\
                    &        &    Male & 13.42\ciformat{0.70} & 13.99\ciformat{1.07} & 13.96\ciformat{0.75} & 13.58\ciformat{0.72} & 13.47\ciformat{0.69} & 13.68\ciformat{0.36} \\
                    &        & Average & 13.23\ciformat{0.48} & 13.95\ciformat{0.76} & 13.52\ciformat{0.50} & 13.29\ciformat{0.49} & 13.36\ciformat{0.49} & 13.47\ciformat{0.25} \\
    
    \textsc{mWhisper-F} &     UK &  Female & 19.37\ciformat{1.04} & 19.08\ciformat{1.05} & 19.27\ciformat{1.06} & 18.79\ciformat{1.00} & 19.04\ciformat{1.02} & 19.11\ciformat{0.46} \\
                    &        &    Male & 19.94\ciformat{1.06} & 20.31\ciformat{1.17} & 21.06\ciformat{1.26} & 20.64\ciformat{1.13} & 20.29\ciformat{1.12} & 20.45\ciformat{0.51} \\
                    &        & Average & 19.65\ciformat{0.74} & 19.70\ciformat{0.79} & 20.16\ciformat{0.82} & 19.72\ciformat{0.76} & 19.66\ciformat{0.76} & 19.78\ciformat{0.35} \\
        \midrule
             \textsc{Gemini} & Indian &  Female & 11.29\ciformat{0.79} & 10.88\ciformat{0.76} & 11.03\ciformat{0.76} & 10.95\ciformat{0.76} & 10.68\ciformat{0.76} & 10.97\ciformat{0.34} \\
                    &        &    Male & 10.52\ciformat{0.71} & 10.86\ciformat{0.78} & 11.30\ciformat{0.78} & 10.79\ciformat{0.75} & 10.11\ciformat{0.69} & 10.72\ciformat{0.33} \\
                    &        & Average & 10.90\ciformat{0.53} & 10.87\ciformat{0.54} & 11.17\ciformat{0.55} & 10.87\ciformat{0.53} & 10.40\ciformat{0.51} & 10.84\ciformat{0.24} \\
    
             \textsc{Gemini} &     US &  Female & 8.33\ciformat{0.53} & 8.34\ciformat{0.54} & 8.58\ciformat{0.56} & 8.34\ciformat{0.56} & 8.34\ciformat{0.55} & 8.39\ciformat{0.25} \\
                    &        &    Male & 8.59\ciformat{0.59} & 8.44\ciformat{0.57} & 8.40\ciformat{0.56} & 8.51\ciformat{0.55} & 8.38\ciformat{0.55} & 8.47\ciformat{0.25} \\
                    &        & Average & 8.46\ciformat{0.40} & 8.39\ciformat{0.39} & 8.49\ciformat{0.39} & 8.42\ciformat{0.39} & 8.36\ciformat{0.39} & 8.43\ciformat{0.18} \\
    
             \textsc{Gemini} &     UK &  Female & 11.70\ciformat{0.87} & 11.19\ciformat{0.86} & 11.96\ciformat{0.88} & 11.42\ciformat{0.84} & 11.53\ciformat{0.85} & 11.56\ciformat{0.38} \\
                    &        &    Male & 11.98\ciformat{0.89} & 11.87\ciformat{0.91} & 12.09\ciformat{0.91} & 11.95\ciformat{0.90} & 11.67\ciformat{0.89} & 11.91\ciformat{0.40} \\
                    &        & Average & 11.84\ciformat{0.62} & 11.53\ciformat{0.63} & 12.02\ciformat{0.63} & 11.69\ciformat{0.62} & 11.60\ciformat{0.61} & 11.73\ciformat{0.28} \\
        \midrule
    \textsc{mWhisper-F} & Indian &  Female & 57.87\ciformat{1.63} & 58.35\ciformat{1.63} & 58.19\ciformat{1.60} & 58.42\ciformat{1.64} & 58.94\ciformat{1.64} & 58.35\ciformat{0.73} \\
                    &        &    Male & 57.62\ciformat{1.57} & 58.04\ciformat{1.62} & 59.33\ciformat{1.71} & 58.33\ciformat{1.80} & 57.98\ciformat{1.61} & 58.26\ciformat{0.74} \\
                    &        & Average & 57.74\ciformat{1.13} & 58.20\ciformat{1.15} & 58.76\ciformat{1.17} & 58.38\ciformat{1.22} & 58.46\ciformat{1.15} & 58.31\ciformat{0.52} \\
    
    \textsc{mWhisper-F} &     US &  Female & 33.28\ciformat{1.08} & 34.81\ciformat{1.21} & 34.07\ciformat{1.13} & 34.58\ciformat{2.10} & 33.40\ciformat{1.10} & 34.03\ciformat{0.62} \\
                    &        &    Male & 34.52\ciformat{1.24} & 35.67\ciformat{1.39} & 37.23\ciformat{1.60} & 34.96\ciformat{1.29} & 35.00\ciformat{1.30} & 35.48\ciformat{0.61} \\
                    &        & Average & 33.90\ciformat{0.82} & 35.24\ciformat{0.92} & 35.65\ciformat{0.98} & 34.77\ciformat{1.23} & 34.20\ciformat{0.85} & 34.75\ciformat{0.43} \\
    
    \textsc{mWhisper-F} &     UK &  Female & 46.78\ciformat{1.58} & 47.41\ciformat{1.65} & 46.00\ciformat{1.57} & 47.07\ciformat{2.25} & 46.36\ciformat{1.56} & 46.73\ciformat{0.78} \\
                    &        &    Male & 45.65\ciformat{1.57} & 46.64\ciformat{1.67} & 47.99\ciformat{1.90} & 45.15\ciformat{1.59} & 45.86\ciformat{1.62} & 46.26\ciformat{0.75} \\
                    &        & Average & 46.21\ciformat{1.11} & 47.03\ciformat{1.17} & 47.00\ciformat{1.23} & 46.11\ciformat{1.38} & 46.11\ciformat{1.13} & 46.49\ciformat{0.54} \\
        \midrule
             \textsc{Gemini} & Indian &  Female & 39.02\ciformat{1.46} & 39.08\ciformat{1.45} & 38.89\ciformat{1.47} & 38.76\ciformat{1.46} & 38.76\ciformat{1.49} & 38.90\ciformat{0.66} \\
                    &        &    Male & 40.39\ciformat{2.58} & 38.87\ciformat{1.48} & 40.29\ciformat{1.56} & 39.84\ciformat{1.86} & 37.66\ciformat{1.45} & 39.41\ciformat{0.82} \\
                    &        & Average & 39.70\ciformat{1.48} & 38.97\ciformat{1.04} & 39.59\ciformat{1.07} & 39.30\ciformat{1.18} & 38.21\ciformat{1.04} & 39.16\ciformat{0.53} \\
    
             \textsc{Gemini} &     US &  Female & 23.74\ciformat{0.96} & 23.69\ciformat{0.97} & 23.64\ciformat{0.96} & 23.97\ciformat{1.00} & 23.47\ciformat{0.95} & 23.70\ciformat{0.43} \\
                    &        &    Male & 23.51\ciformat{0.95} & 23.85\ciformat{0.97} & 23.56\ciformat{1.04} & 23.56\ciformat{0.95} & 23.02\ciformat{0.91} & 23.50\ciformat{0.43} \\
                    &        & Average & 23.62\ciformat{0.68} & 23.77\ciformat{0.68} & 23.60\ciformat{0.71} & 23.76\ciformat{0.69} & 23.25\ciformat{0.66} & 23.60\ciformat{0.31} \\
    
             \textsc{Gemini} &     UK &  Female & 34.30\ciformat{1.56} & 34.31\ciformat{1.54} & 34.00\ciformat{1.47} & 34.07\ciformat{1.47} & 33.67\ciformat{1.48} & 34.07\ciformat{0.67} \\
                    &        &    Male & 34.20\ciformat{1.49} & 35.30\ciformat{1.80} & 34.71\ciformat{1.53} & 34.42\ciformat{1.53} & 33.87\ciformat{1.53} & 34.50\ciformat{0.71} \\
                    &        & Average & 34.25\ciformat{1.08} & 34.81\ciformat{1.18} & 34.36\ciformat{1.06} & 34.24\ciformat{1.06} & 33.77\ciformat{1.06} & 34.29\ciformat{0.49} \\
        \bottomrule
    \end{tabular}
    \caption{All WER results with confidence intervals across all splits.}
    \label{tab:all-results}
\end{table*}

\begin{table*}[t]
    \begin{tabular}{lp{0.8\textwidth}}
        \toprule
        Reference & Each Bantu group changes musical style with every change of language. \\
        \midrule
        Hypothesis & This is my, okay. My name is Olanrewaja. I just want to tell you about the future. The future about getting crypto to convert cash to crypto. Okay, a lot of people like converting their crypto into cash and they don't want to convert it like through any exchange or any, okay, anything. Because some exchanges have very high prices and some exchanges have very low prices, so a lot of people just want to get to someone that will convert their crypto and convert it into cash and give it to them, or convert the cash to crypto, and they'll send it to them. That's a good business because you can earn on this, a lot. Thank you. \\
        \midrule
        Reference & Potatoes are less space-saving, durable and cheap than pasta. \\
        \midrule
        Hypothesis & Cheetos R S P C E D N J R $1$ $2$ $3$ N S M B Q R S. \\
        \bottomrule
    \end{tabular}
    \caption{\textsc{Gemini-$2.5$-flash} hallucinations.}
    \label{tab:gemini-hallucinations}
\end{table*}

Sometimes \textsc{Gemini} refused to transcribe part of the input, replacing parts of the text with some variation of [unintelligible]. Table \ref{tab:gemini-partial-refusal} provides an example.

\begin{table*}[t]
    \begin{tabular}{lp{0.8\textwidth}}
        \toprule
        Reference & Historians and Ma Chao's contemporaries have a generally negative view of him. \\
        \midrule
        Hypothesis & Historians have [unintelligible] contemporaries [unintelligible] a [unintelligible] negative view of [unintelligible]. \\
        \bottomrule
    \end{tabular}
    \caption{\textsc{Gemini-$2.5$-flash} partial refusals.}
    \label{tab:gemini-partial-refusal}
\end{table*}

\textsc{Gemini} also refused to transcribe some audios in full. Table \ref{tab:gemini-full-refusal} provides some examples. Sometimes, these descriptions lead to a longer output than the reference text, resulting in a WER higher than 100.

\begin{table*}[t]
    \begin{tabular}{lp{0.8\textwidth}}
        \toprule
        Short description & ``[No discernible speech from the main speaker. Background chatter is present.]'', ``[no audible speech]'', ``[Muffled voices and background chatter]'', ``[No discernible speech]'', ``[Indistinct chatter]'' \\
        \midrule
        Long description & ``[The main speaker's speech is unintelligible due to overwhelming background noise.]'',  ``The audio quality for the main speaker is very poor, making it difficult to isolate words clearly due to significant background chatter. The main speaker's speech is almost entirely unintelligible, consisting mostly of mumbles or very soft sounds that are drowned out. Therefore, I cannot accurately transcribe what the main speaker is saying.''\\
        \bottomrule
    \end{tabular}
    \caption{\textsc{Gemini-$2.5$-flash} full refusals.}
    \label{tab:gemini-full-refusal}
\end{table*}

Table \ref{tab:gemini-sound-descriptions} provides examples of \textsc{Gemini} adding sound descriptions on top of the transcribed text.

\begin{table*}[t]
    \begin{tabular}{lp{0.8\textwidth}}
        \toprule
        Reference & Chickens lay so many eggs a week, it wouldn't be possible to breed them all. \\
        \midrule
        Hypothesis & [Sound of firework/firecracker] Chickens lay so many eggs a week, it wouldn't be possible to breed them all. \\
         & [Loud bang] Chickens lay so many eggs a week, it wouldn't be possible to breed them all. \\
         & [EXPLOSION] Chickens lay so many eggs a week, it wouldn't be possible to breed them all. \\
         & [Sound of fireworks/gunshot] Chickens lay so many eggs a week, it wouldn't be possible to breed them all. \\
        \bottomrule
    \end{tabular}
    \caption{\textsc{Gemini-$2.5$-flash} sound descriptions.}
    \label{tab:gemini-sound-descriptions}
\end{table*}

Table \ref{tab:gemini-repeat-prompt} provides an example of \textsc{Gemini} repeating part of the prompt in the output.

\begin{table*}[t]
    \begin{tabular}{lp{0.8\textwidth}}
        \toprule
        Reference & He also won the South African National Time Trial Championships at junior level. \\
        \hline
        Hypothesis & You on the all right. You on the all right to go out there when you saw that happen on the field, and that was just an unbelievable moment, and I think a very emotional one for that whole team.The user wants a transcription of the audio. I will listen carefully to the speaker's words and transcribe them, focusing on accuracy and natural speech flow. I need to be aware of any background noise and try to isolate the main speaker. \\
        \bottomrule
    \end{tabular}
    \caption{\textsc{Gemini-$2.5$-flash} prompt repetition.}
    \label{tab:gemini-repeat-prompt}
\end{table*}

Finally, Table \ref{tab:gemini-bad-refusals} provides an example of an unacceptable refusal. 

\begin{table*}[t]
    \begin{tabular}{lp{0.8\textwidth}}
        \toprule
        Reference & Historians and Ma Chao's contemporaries have a generally negative view of him. \\
        \hline
        Hypothesis & The audio appears to be in Spanish. The speaker says: "Historia con el más chachi, contemporánea, apagándole la vida a la gente que y uno va a". This translates to: "History with the coolest, contemporary, turning off the lives of people who and one goes to." Please confirm if you would still like this to be transcribed in English, or if you meant a transcription of the Spanish spoken. \\
        \bottomrule
    \end{tabular}
    \caption{\textsc{Gemini-$2.5$-flash} unacceptable refusal.}
    \label{tab:gemini-bad-refusals}
\end{table*}

\section{Compute}\label{app:Compute}
For our experiments with the \textsc{Whisper-Flamingo} model family, we used two NVIDIA RTX A$6000$ GPUs for approximately $500$ hours.
For our experiments with \textsc{Gemini-$2.5$-flash}, we used approximately $1$B input and output tokens and an Intel(R) Xeon(R) CPU E$5$-$2673$ v$3$ @ $2.40$GHz for approximately $400$ hours.

\section{Use of AI Assistants}\label{app:AI_assist}
We used Deepseek to look up statistical analysis references, which were read, corroborated, and verified by hand. We also used Copilot and Genie for coding support, which was also manually checked.

\end{document}